\documentclass[preprint,12pt]{elsarticle}

\usepackage{amsmath,amssymb}
\usepackage[linesnumbered,ruled,vlined]{algorithm2e}
\usepackage{lineno}

\begin{document}

\begin{frontmatter}

\title{Real-Time Neural Network Scheduling of Emergency Medical Mask Production during COVID-19}


\author[hznu-hise]{Chen-Xin Wu}
\author[hznu-hise]{Min-Hui Liao}
\author[ndu]{Mumtaz Karatas}
\author[tjut]{Sheng-Yong Chen}
\author[hznu-hise]{Yu-Jun~Zheng\corref{cor}}
\cortext[cor]{Corresponding author.}
\ead{yujun.zheng@computer.org}

\address[hznu-hise]{College of Information Science and Engineering, Hangzhou Normal University, Hangzhou 311121, China}
\address[ndu]{Industrial Engineering Department, Naval Academy, National Defense University, Tuzla 34940, Istanbul, Turkey}
\address[tjut]{School of Computer Science and Engineering, Tianjin University of Technology, Tianjin 300384, China.}

\begin{abstract}
During the outbreak of the novel coronavirus pneumonia (COVID-19), there is a huge demand for medical masks. A mask manufacturer often receives a large amount of orders that are beyond its capability. Therefore, it is of critical importance for the manufacturer to schedule mask production tasks as efficiently as possible. However, existing scheduling methods typically require a considerable amount of computational resources and, therefore, cannot effectively cope with the surge of orders. In this paper, we propose an end-to-end neural network for scheduling real-time production tasks. The neural network takes a sequence of production tasks as inputs to predict a distribution over different schedules, employs reinforcement learning to optimize network parameters using the negative total tardiness as the reward signal, and finally produces a high-quality solution to the scheduling problem. We applied the proposed approach to schedule emergency production tasks for a medical mask manufacturer during the peak of COVID-19 in China. Computational results show that the neural network scheduler can solve problem instances with hundreds of tasks within seconds. The objective function value (i.e., the total weighted tardiness) produced by the neural network scheduler is significantly better than those of existing constructive heuristics, and is very close to those of the state-of-the-art metaheuristics whose computational time is unaffordable in practice.
\end{abstract}



\begin{keyword}
emergency production\sep flow shop scheduling\sep neural network\sep reinforcement learning\sep public health emergencies
\end{keyword}

\end{frontmatter}


\section{Introduction}\label{sec:intr}
ZHENDE is a medical apparatus manufacturer in Zhejiang Province, China. It has a mask production line that can produce different types masks, such as disposable medical masks, surgical masks, medical protective masks, and respiratory masks. The Daily output is nearly one hundred thousand. However, on each day during the outbreak of the novel coronavirus pneumonia (COVID-19), the manufacturer often receives tens to hundreds of mask orders, the total demand of which ranges from hundreds of thousands to a million masks. Almost all orders have tight delivery deadlines. Therefore, it is of critical importance for the manufacturer to efficiently schedule the mask production tasks to satisfy the orders as much as possible. The manufacturer asked our research team to develop a production scheduler that can schedule hundreds of tasks within seconds. In fact, many manufacturers of medical supplies have similar requirements during the pandemic.

Scheduling production tasks on a production line can be formulated as a machine scheduling problem which is known to be NP-hard \cite{Pinedo02Scheduling}. Exact optimization algorithms (e.g., \cite{McMahon67,Karlof96COR,Ziaee07AMC,Gicquel12COR}) have very large computation times that are infeasible on even moderate-size problem instances. As for moderate- and large-size instances optimal solutions are rarely needed in practice, heuristic approximation algorithms, in particular evolutionary algorithms (e.g., \cite{Etiler04JORS,Onwu06EJOR,Liao07CAOR,Kuo09ESA,Lin16IJPR,Zhao18ESA,Zheng19ASOC}), are more feasible to achieve a trade-off between optimality and computational costs. However, the number of repeated generations and objective function evaluations for solving large-size instances still takes a relatively long time and, therefore, cannot satisfy the requirement of real-time scheduling.

Using end-to-end neural networks to directly map a problem input to an optimal or near-optimal solution is another research direction that has received increasing attention. The earliest work dates back to Hopfield and Tank \cite{Hopfield85}, who applied a Hopfield-network to solve the traveling salesman problem (TSP). Simon and Takefuji \cite{Simon88} modified the Hopfield network to solve the job-shop scheduling problem. However, the Hopfield network is only suitable for very small problem instances. Based on the premise that optimal solutions to a scheduling problem have common features which can be implicitly captured by machine learning, Weckman et al. \cite{Weck08JIM} proposed a neural network for scheduling job-shops by capturing the predictive knowledge regarding the assignment of operation's position in a sequence. They used solutions obtained by genetic algorithm (GA) as samples for training the network. To solve the flow shop scheduling problem, Ramanan et al. \cite{Rama11JIM} used a neural network trained with optimal solutions of known instances to produce quality solutions for new instances, which are then given as the initial solutions to improve other heuristics such as GA. Recently, deep learning has been utilized to optimization algorithm design by learning algorithmic decisions based on the distribution of problem instances. Vinyals et al. \cite{Vinyals15NIPS} introduced the pointer network as a sequence-to-sequence model, which consists in an encoder to parse the input nodes, and a decoder to produce a probability distribution over these nodes based on a pointer (attention) mechanism over the encoded nodes. They applied the pointer network to solve TSP instances with up to 100 nodes. However, the pointer network is trained in a supervised manner, which heavily relies on the expensive optimal solutions of sample instances. Nazari et al. \cite{Nazari18NIPS} addressed this difficulty by introducing reinforcement learning to calculate the rewards of output solutions, and applied the model to solve the vehicle routing problem (VRP). Kool et al. \cite{Kool19ICLR} used a different decoder based on a context vector and improved the training algorithm based on a greedy rollout baseline. They applied the model to several combinatorial optimization problems including TSP and VRP. Peng et al. \cite{Peng20ISICA} presented a dynamic attention model with dynamic encoder-decoder architecture to exploit hidden structure information at different construction steps, so as to construct better solutions. 

In this paper, we propose a deep reinforcement approach for scheduling real-time production tasks. The neural network takes a sequence of production tasks as inputs to predict a distribution over different schedules, employs reinforcement learning to optimize network parameters using the negative total tardiness as the reward signal, and finally produces a high-quality task scheduling solution. We applied the proposed neural network scheduler to a medical mask manufacturer during the peak of COVID-19 in China. Computational results show that the neural network scheduler can solve problem instances with hundreds of tasks within seconds. The objective function value (i.e., the total weighted tardiness) produced by the neural network scheduler is significantly better than those of existing constructive heuristics such as the Nawaz, Enscore and Ham (NEH) heuristic \cite{Nawaz83} and Suliman heuristic \cite{Suliman00IJPE}, and is very close to those of the state-of-the-art metaheuristics whose computational time is obviously unaffordable in practice.

The remainder of this paper is organized as follows. Section 2 describes the emergency production scheduling problem. Section 3 presents the architecture of the neural network, Section 4 depicts the reinforcement learning algorithm, and Section 5 presents the experimental results, and finally Section 6 concludes with a discussion.

\section{Medical Mask Production Scheduling Problem}\label{sec:prob}
In this section, we formulate the scheduling problem as follows (the variables are listed in Table \ref{tab:nom}). The manufacturer receives $K$ orders, denoted by $\mathbf{O}=\{O_1,O_2,\dots,O_K\}$. Each order $O_k$ is associated with a set $\Phi_k$ of production tasks (jobs), which is related to the number of mask types in the order. Each order $O_k$ has an expected delivery time $d_k$ and an importance weight $w_k$ according to its value and urgency. In our practice, the manager gives a score between 1--10 for each order, and then all weights are normalized such that $(\sum_{k=1}^Kw_k)=1$.

Let $\mathbf{J}=\{J_1,J_2,\dots,J_n\}$ be the set of all tasks. These tasks need to be scheduled on a production line with $m$ machines, denoted by $\mathbf{M}=\{M_1,M_2,\dots,M_m\}$. Each task $J_j$ has exactly $m$ operations, where the $i$-th operation must be processed on machine $M_i$ with a processing time $t_{ij}$ ($1\le i\le m;1\le j\le n$). Each machine can process at most one task at a time, and each operation cannot be interrupted. The operations of mask production typically include cloth cutting, fabric lamination, belt welding, disinfection, and packaging.

\begin{table}
\center\small\caption{Mathematical variables used in the problem formulation.}
\begin{tabular}{c|l}\hline
Symbol & Description\\\hline
$\mathbf{O}=\{O_1,O_2,\dots,O_K\}$ & The set of orders\\
$K$ & Number of orders\\
$k$ & Index of orders ($1\!\le\!k\!\le\!K$)\\
$d_k$ & Expected delivery time of order $O_k$\\
$w_k$ & Importance weight of order $O_k$\\
$\Phi_k$ & Set of all production tasks in order $O_k$\\
$\mathbf{J}=\{J_1,J_2,\dots,J_n\}$ & Set of all production tasks\\
$n$ & Number of tasks\\
$j$ & Index of tasks ($1\!\le\!j\!\le\!n$)\\
$\mathbf{M}=\{M_1,M_2,\dots,M_m\}$ & Set of machines\\
$m$ & Number of machines\\
$i$ & Index of machines ($1\!\le\!i\!\le\!m$)\\
$t_{ij}$ & Processing time of $i$-th operation of task $J_j$ (on machine $M_i$)\\
$\boldsymbol{\pi}=\{\pi_1,\pi_2,\dots,\pi_n\}$ & A solution (sequence of $n$ tasks) to the problem\\
$C(\pi_{i},j)$ & Completion time of task $\pi_j$ on machine $M_i$\\
$T(O_k)$ & Completion time of order $O_k$ \\\hline
\end{tabular}\label{tab:nom}
\end{table}

The problem is to decide a processing sequence $\boldsymbol{\pi}=\{\pi_1,\pi_2,\dots,\pi_n\}$ of the $n$ tasks. Let $C(\pi_{i},j)$ denote the completion time of task $\pi_j$ on machine $M_i$. For the first machine $M_1$, the tasks can be sequentially processed immediately one by one:
\begin{eqnarray}
C(\pi_1,1)&=& t_{\pi_1,1} \label{eq:mach11}\\
C(\pi_j,1)&=& C(\pi_{j-1},1)+t_{\pi_j,1}, \quad j=2,...,n \label{eq:mach1j}
\end{eqnarray}

The first job $\pi_1$ can be processed on each subsequent machine $M_i$ immediately after it is completed on the previous machine $M_{i-1}$:
\begin{equation}
C(\pi_1,i)=C(\pi_1,i-1)+t_{\pi_1,i}, \quad i=2,...,m
\label{eq:job1}\end{equation}

Each subsequent job $\pi_j$ can be processed on machine $M_i$ only when (1) the job $\pi_j$ has been completed on the previous machine $M_{i-1}$; (2) the previous job $\pi_{j-1}$ has been completed on machine $M_i$:
\begin{equation}
C(\pi_j,i)=\max\big(C(\pi_j,i-1),C(\pi_{j-1},i)\big)+t_{\pi_j,i}, \quad i=2,\dots,m; j=2,\dots,n
\label{eq:jobj}\end{equation}

Therefore, the completion time of each order $O_k$ is the completion time of the last task of the order on machine $M_m$:
\begin{equation}
T(O_k)=\max_{\pi\in \Phi_k}C(\pi,m)
\label{eq:makespan}\end{equation}

The objective of the problem is to minimize the total weighted tardiness of the orders:
\begin{equation}
\min f(\boldsymbol{\pi})=\sum_{k=1}^{K} w_k\max(T(O_k)-d_k,0)
\label{eq:obj}\end{equation}

If all tasks are available for processing at time zero, the above formulation can be regarded as a variant of the permutation flow shop scheduling problem which is known to be NP-hard \cite{Pinedo02Scheduling}. When there are hundreds of tasks to be scheduled, the problem instances are computationally intractable for exact optimization algorithms, and search-based heuristics also typically take tens of minutes to hours to obtain a satisfying solution. Moreover, in a public health emergency such as the COVID-19 pandemic, new orders may continually arrive during the emergency production and, therefore, it is frequently to reschedule  production tasks to incorporate new tasks into the schedules. The allowable computational time for rescheduling is even shorter, typical only a few seconds. Hence, it is required to design real-time or near-real-time rescheduling methods for the problem.

\section{A Neural Network Scheduler for Emergence Production Task Scheduling}
\begin{figure}
\centering
\includegraphics[scale=0.83]{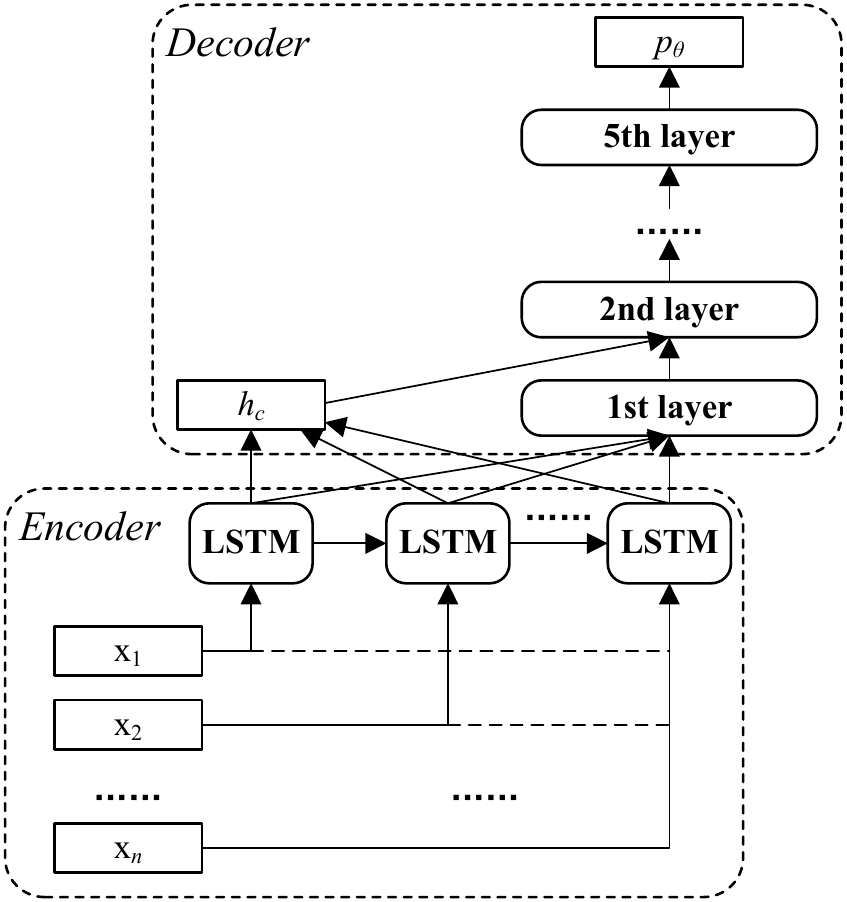}
\caption{Architecture of the neural network scheduler.}
\label{fig:arch}\end{figure}

We propose a neural network scheduler based on the encoder-decoder architecture \cite{Sutskever14NIPS} to efficiently solve the above production task scheduling problem. Fig. \ref{fig:arch} illustrates the architecture of the network. The input to the network is a problem instance represented by a sequence of $n$ tasks, each of which is described by a $(m\!+\!2)$-dimensional vector $\mathbf{x}_j=\{p_{j,1},p_{j,2},\dots,p_{j,m},d_k,w_k\}$ that consists the processing times on the $m$ machines and the expected delivery time and weight importance of the corresponding order. To facilitate the processing of the neural network, all inputs are normalized into [0,1], e.g., each $d_k$ is transformed to $(d_k\!-\!d_{\min})/(d_{\max}\!-\!d_{\min})$, where $d_{\min}=\min_{1\!\le\!k\!\le\!K}d_k$ and $d_{\max}=\max_{1\!\le\!k\!\le\!K}d_k$.

The encoder is a recurrent neural networks (RNN) with long short-term memory (LSTM) \cite{Gers99IET} cells. An LSTM takes a task $\mathbf{x}_j$ as input at a time and transforms it to a hidden state $\mathbf{h}_j$ by increasingly computing the embedding of the inputs (where $\textit{att}$ denotes the transformation by LSTM):
\begin{eqnarray}
\mathbf{h}_1&=& \textit{att}(\mathbf{h}_1,\mathbf{x}_1)= \textit{encode}(\mathbf{x}_1) \nonumber\\
\mathbf{h}_2&=& \textit{att}(\mathbf{h}_1,\mathbf{x}_2)= \textit{encode}(\mathbf{x}_1,\mathbf{x}_2) \nonumber\\
\vdots \nonumber\\
\mathbf{h}_n&=& \textit{att}(\mathbf{h}_{n-1},\mathbf{x}_2)= \textit{encode}(\mathbf{x}_1,\mathbf{x}_2,\ldots,\mathbf{x}_n)
\label{eq:encode}\end{eqnarray}

As a result, the encode produces an aggregated embedding of all inputs as the mean of $n$ hidden states:
\begin{equation}
\overline{\mathbf{h}}= \frac{1}{n}\sum_{j=1}^{n}\mathbf{h}_j
\label{eq:mean}\end{equation}

The decoder also performs $n$ decoding steps, each making a decision on which task should to be processed at the next step. At each $j$-th step, it constructs a context vector $\mathbf{h}_c$ by concatenating $\overline{\mathbf{h}}$ and the hidden state $\mathbf{h}_{j-1}$ of the previous LSTM. We use a five-layer deep neural network to implement the decoder. The first layer takes $\overline{\mathbf{h}}$ as input and transforms it into a $n_1$-dimensional hidden vector $\mathbf{u}_1$ ($n_1<n$):
\begin{equation}
\mathbf{u}_1= \mathrm{ReLU}(W_1\overline{\mathbf{h}}^\mathrm{T}+\mathbf{b}_1)
\label{eq:dnn1}\end{equation}
where $W_1$ is a $n_1\!\times\!n$ weight matrix and $\mathbf{b}_1$ is a $n_1$-dimensional bias vector.

The second layer takes the concatenation of $\mathbf{u}_1$ and context vector $\mathbf{h}_c$ as input and transforms it into a $n_2$-dimensional hidden vector $\mathbf{u}_2$:
\begin{equation}
\mathbf{u}_2= \mathrm{ReLU}(W_2[\mathbf{u}_1;\mathbf{h}_c]^\mathrm{T}+\mathbf{b}_2)
\label{eq:dnn2}\end{equation}
where $[\_;\_]$ denotes the horizontal concatenation of vectors, $W_2$ is a $n_2\!\times\!(n_1\!+\!2n)$ weight matrix and $\mathbf{b}_2$ is a $n_2$-dimensional bias vector.

And each of the remaining layer takes the hidden state of the previous layer, and transforms it into a lower-dimensional hidden vector using ReLU activation. Finally, the probability that each task $x_j$ is selected at the $t$-th step is calculated based on the state $\mathbf{u}$ of the top layer of the DNN:
\begin{equation}
p_{\theta}(\pi_t\!=\!x_j|\mathbf{x},\boldsymbol{\pi}_{1:t\!-\!1})= \frac{e^{u_j}}{\sum_{j'=1}^{n}e^{u_{j'}}}
\label{eq:decode}\end{equation}

\section{Reinforcement Learning of the Neural Network}
A solution to the scheduling problem can be viewed as a sequence of decisions, and the decision process can be regarded as a Markov decision process \cite{Bello16arXiv}. According to the objective function (\ref{eq:obj}), the training of the network is to minimize the loss
\begin{equation}
\mathcal{L}(\boldsymbol{\theta}|\mathbf{x})= \mathbb{E}_{\boldsymbol{\pi}\sim p_{\theta}(\_|\mathbf{x})}f(\boldsymbol{\pi}|\mathbf{x})
\label{eq:loss}\end{equation}

We employ the policy gradient using REINFORCE algorithm \cite{Williams92ML} with Adam optimizer \cite{King15Adam} to train the network. The gradients of network parameters $\boldsymbol{\theta}$ are defined based on a baseline $\textit{base}(\mathbf{x})$ as:
\begin{equation}
\nabla_{\boldsymbol{\theta}}\mathcal{L}(\boldsymbol{\theta}|\mathbf{x})= \mathbb{E}_{\boldsymbol{\pi}\sim p_{\theta}(\_|\mathbf{x})}\big((f(\boldsymbol{\pi}|\mathbf{x})\!-\!\textit{base}(\mathbf{x}))\nabla_{\boldsymbol{\theta}}\log p_{\theta}(\boldsymbol{\pi}|\mathbf{x})\big)
\label{eq:gradient}\end{equation}

A good baseline reduces gradient variance and increases learning speed \cite{Kool19ICLR}. Here, we use both the NEH heuristic \cite{Nawaz83} and Suliman heuristic \cite{Suliman00IJPE} to solve each instance $\mathbf{x}$, and use the better one as the $\textit{base}(\mathbf{x})$.

During the training, we approximate the gradient via Monte-Carlo sampling, where $B$ problem instances are drawn from the same distribution:
\begin{equation}
\nabla_{\boldsymbol{\theta}}\mathcal{L}(\boldsymbol{\theta})= \frac{1}{B}\sum_{i=1}^{B}\big((f(\boldsymbol{\pi}_i|\mathbf{x}_i)\!-\!\textit{base}(\mathbf{x}_i))\nabla_{\boldsymbol{\theta}}\log p_{\theta}(\boldsymbol{\pi}_i|\mathbf{x}_i)\big)
\label{eq:gradient}\end{equation}

Algorithm \ref{alg:rein} presents the pseudocode of the REINFORCE algorithm.

\begin{algorithm}\small
Randomly initialize the network parameters $\boldsymbol{\theta}$\;
\For{epoch$=1$ to epoch$_{\max}$}{
    \For{$t=1$ to $T$}{
        \For{$i=1$ to $B$}{
            Sample an instance $\mathbf{x}_i$ from the problem distribution\;
            Compute the model output $\boldsymbol{\pi}_i$\;
            Compute the baseline $\textit{base}(\mathbf{x}_i)$\;
        }
        $g(\boldsymbol{\theta})\gets \frac{1}{B}\sum_{i=1}^{B}\big((f(\boldsymbol{\pi}_i)\!-\!\textit{base}(\mathbf{x}_i))\nabla_{\boldsymbol{\theta}}\log p_{\theta}(\boldsymbol{\pi}_i|\mathbf{x}_i)\big)$\;
        $\boldsymbol{\theta}\gets \mathrm{Adam}(\boldsymbol{\theta},g(\boldsymbol{\theta}))$\;
    }
}
\Return{$\boldsymbol{\theta}$.}
\caption{The REINFORCE algorithm.}
\label{alg:rein}\end{algorithm}

\section{Computational Results}
In the training phase, according to production tasks of the manufacturer during the peak of COVID-19 in China, we randomly generate 20,000 instances. The basic features of the instance distribution are as follows: $m\!=\!5$, $n$ follows a normal distribution $N(124,33)$, $t_{ij}$ follows a normal distribution $N(2.4,1.6)$ (in hours), and $d_k$ follows a uniform discrete distribution $\{24,36,48,60,72,96,120\}$ (in hours). The maximum number of epochs for training the network is set to 100.

For comparison, we also use three different baselines: the first is a greedy heuristic that sorts tasks in decreasing order of $w_k/(\sum_{i=1}^{m}t_{ij})$, the second is the NEH heuristic \cite{Nawaz83}, and the third is the Suliman heuristic \cite{Suliman00IJPE}. The neural network is implemented using Python 3.4, while the heuristics are implemented with Microsoft Visual C\# 2018. The experiments are conducted on a computer with Intel Xeon 3430 CPU and GeForce GTX 1080Ti GPU.

\begin{figure}
\centering
\includegraphics[scale=0.63]{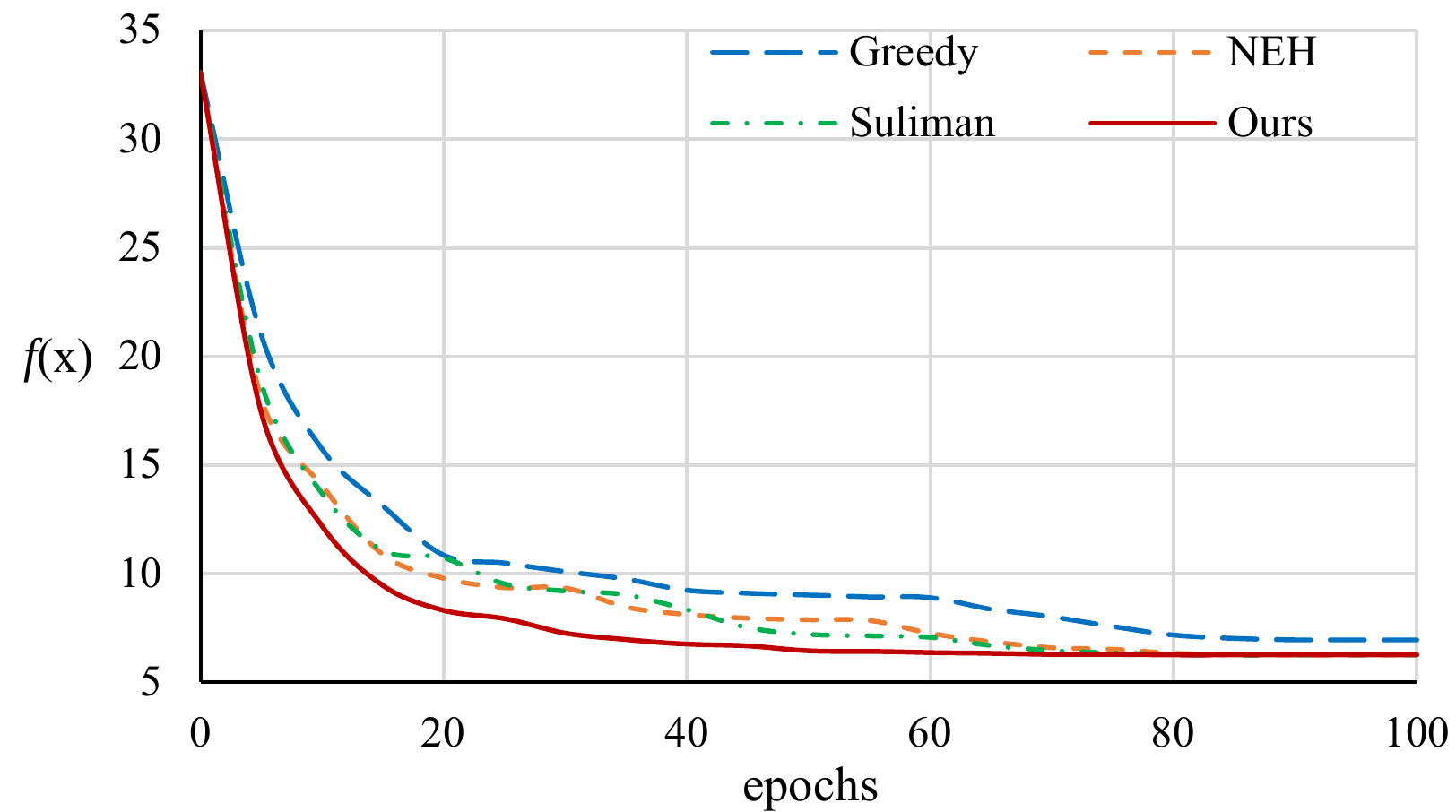}
\caption{The convergence curves of the four methods for training the neural network.}
\label{fig:train}\end{figure}

Fig. \ref{fig:train} presents the convergence curves of the four methods during the training process. In average, our method converges after 55$\sim$65 epochs, the individual NEH and Suliman heuristics converge after 80$\sim$85 epochs, while the greedy method converges to local optima that are significantly worse than the results of the first three methods. The results demonstrate that our method using hybrid NEH and Suliman heuristics as the baseline can significantly improve the training performance compare to the existing heuristic baselines.

Next, we test the performance of the trained neural network scheduler for solving the mask production task scheduling problem. We select 146 real-world instances of the manufacturer from Feb 8 to Feb 14, 2020, the peak of COVID-19 in China. For each day, we need to first solve an instance with about 50$\sim$200 tasks; during the daytime, with the arrival of new orders, we need to reschedule the production for 20$\sim$40 times. To validate the performance of neural network scheduler, we also run the following five state-of-the-art metaheuristic algorithms to solve each instance, as use the best result among the algorithms as the benchmark:
\begin{itemize}
\item A shuffled complex evolution algorithm (SCEA) \cite{Zhao15IJCIM};
\item An algebraic differential evolution (ADE) algorithm \cite{Sant16TEVC};
\item A teaching-learning based optimization (TLBO) algorithm \cite{Shao17ASOC};
\item A biogeography-based optimization (BBO) algorithm \cite{Du18BICTA};
\item A discrete water wave optimization (WWO) algorithm \cite{Zheng19ASOC}.
\end{itemize}

\begin{figure}
\centering
\includegraphics[scale=0.72]{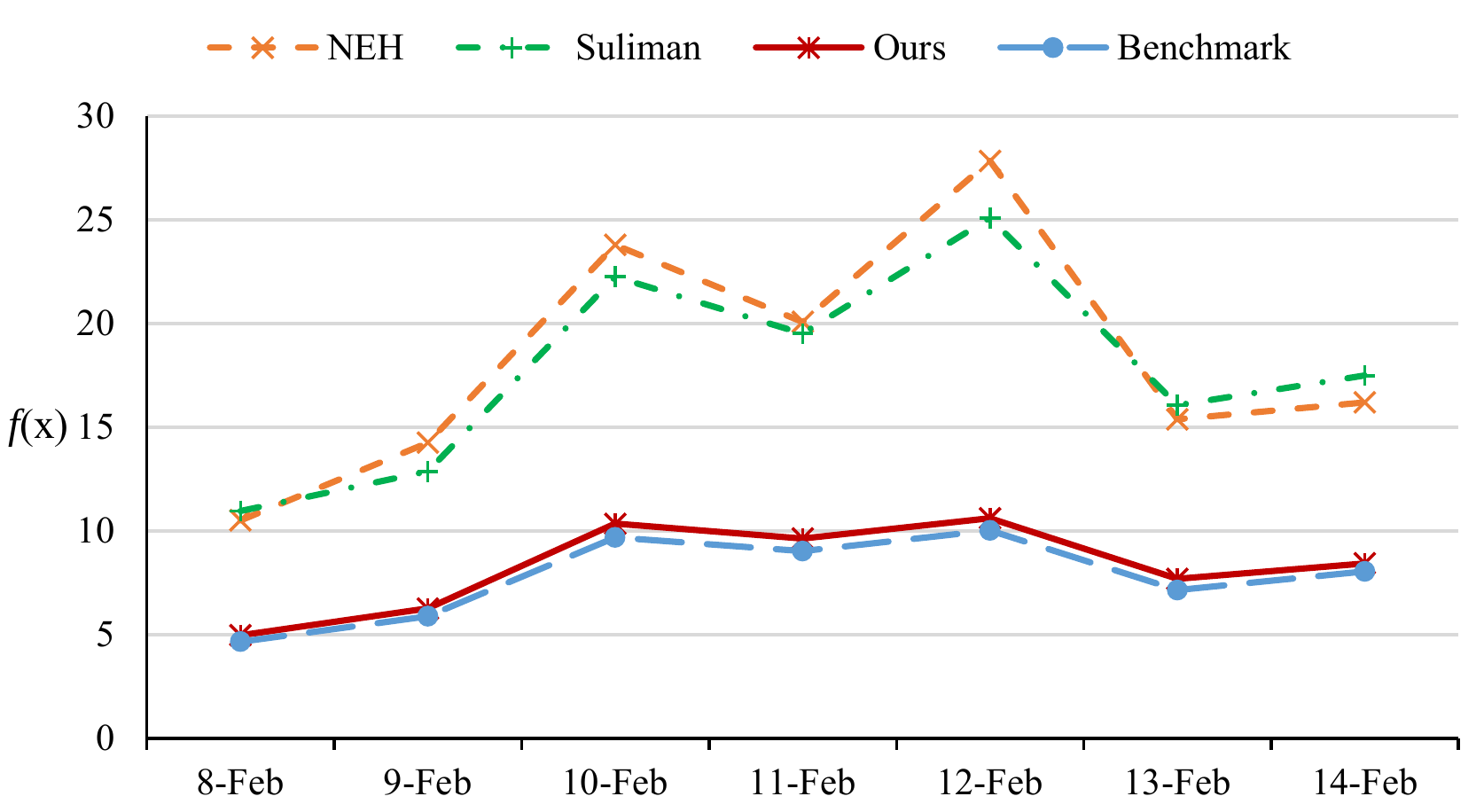}
\caption{Comparison of the results of the neural network scheduler, constructive heuristics, and state-of-the-art metaheuristics.}
\label{fig:app}\end{figure}

Fig. \ref{fig:app} presents the average objective function value obtained by our scheduler and those of the NEH, Suliman, and benchmark solutions on each day, and Table \ref{tab:CT} presents the average CPU time required to obtain the solutions. The results show that, the results of the neural network scheduler are significantly better than those of the NEH and Suliman heuristics. The NEH heuristic and neural network scheduler consume similar computational time, but the objective function value of NEH is about 2$\sim$3 times of that of the neural network scheduler. The Suliman heuristic consumes more computational time and obtains even worse objective function value than the neural network scheduler. The benchmark solutions are obtained by the best metaheuristic among the five state-of-the-art ones, using significantly longer computational time (600$\sim$1500 seconds) than that of the neural network scheduler (only 1$\sim$2 seconds). Nevertheless, the objective function values produced by the neural network scheduler are very close to (approximately 6\%$\sim$7\% larger than) those of the benchmark solutions. In emergency conditions, the computational time of the state-of-the-art metaheuristics is obviously unaffordable, while the proposed neural network scheduler can produce high-quality solutions within seconds and, therefore, satisfy the requirements of emergency medical mask production.

\begin{table}
\center\caption{CPU time (in seconds) consumed by the neural network scheduler, constructive heuristics, and state-of-the-art metaheuristics.}
\begin{tabular}{l|rrrr}\hline
\multicolumn{1}{c|}{Day} &\multicolumn{1}{c}{NEH} &\multicolumn{1}{c}{Suliman} &\multicolumn{1}{c}{Benchmark algorithm} &\multicolumn{1}{c}{Neural network}\\\hline
Feb-8 &0.63 &0.93 &522 &0.75 \\
Feb-9 &0.92 &2.35 &713 &1.03 \\
Feb-10&1.71 &4.59 &1064 &1.32 \\
Feb-11&2.13 &5.26 &1125 &1.57 \\
Feb-12&2.30 &5.96 &1390 &1.72 \\
Feb-13&0.97 &2.64 &885 &1.09 \\
Feb-14&1.45 &3.84 &971 &1.20 \\\hline
\end{tabular}
\label{tab:CT}\end{table}

\section{Conclusion}
In this paper, we propose a deep neural network with reinforcement for scheduling emergency production tasks within seconds. The neural network consists of an encoder and a decoder. The encoder employs LSTM-based RNN to sequentially parse the input production tasks, and the decoder employs a deep neural network to learn the probability distribution over these tasks. The network is trained by reinforcement learning using the negative total tardiness as the reward signal. We applied the proposed neural network scheduler to a medical mask manufacturer during the peak of COVID-19 in China. The results show that the proposed approach can achieve high-quality solutions within very shorter computational time to satisfy the requirements of emergency production.

The baseline plays a key role in reinforcement learning. The baseline used in this paper is based on two constructive heuristics, which have much room to be improved. However, better heuristics and metaheuristics often require large computational resource and are not efficient in training a large number of test instances. In our ongoing work, we are incorporating other neural network schedulers to improve the baseline. Another future work is to use evolutionary metaheuristics to optimize the parameters of the deep neural network \cite{Zhou19SWEVO}.

\section*{Acknowledgment}
This work was supported by National Natural Science Foundation of China under Grant 61872123 and Zhejiang Provincial Natural Science Foundation under Grant LY18F030023 and LR20F030002.


\end{document}